\PassOptionsToPackage{table}{xcolor}
\documentclass{article} % For LaTeX2e
\usepackage{iclr2027_conference,times}

\usepackage{amsmath,amsfonts,bm}

\def\eqref#1{equation~\ref{#1}}
\def\Eqref#1{Equation~\ref{#1}}
\def\1{\bm{1}}

\DeclareMathAlphabet{\mathsfit}{\encodingdefault}{\sfdefault}{m}{sl}
\SetMathAlphabet{\mathsfit}{bold}{\encodingdefault}{\sfdefault}{bx}{n}

\usepackage{hyperref}
\usepackage{url}
\usepackage{booktabs}
\usepackage{graphicx}
\usepackage{wrapfig}
\usepackage{xcolor}
\usepackage{xcolor,colortbl}
\usepackage{makecell}
\usepackage{amsmath,booktabs,multirow,graphicx}

\definecolor{TableHeader}{RGB}{229,235,240}
\definecolor{TableSubheader}{RGB}{246,248,250}
\definecolor{ReferenceShade}{RGB}{247,248,249}
\definecolor{StrictShade}{RGB}{237,246,242}
\definecolor{BestShade}{RGB}{211,234,221}
\definecolor{SummaryShade}{RGB}{238,243,247}
\definecolor{TopKShade}{RGB}{228,241,234}
\definecolor{MutedTableText}{RGB}{105,113,122}

\title{Rethinking Cross-Tokenizer On-Policy Distillation: From Alignment Coverage to Supervision Reliability}

\author{Bingxi Hou\footnotemark[1]\thanks{Equal Contribution.}, 
Guochao Jiang\rlap{$^*$}, 
Guofeng Quan, 
Weiqing Li, 
Wenfeng Feng, \\
\textbf{Guohua Liu}, 
\textbf{Yuewei Zhang}\footnotemark[2]\thanks{Corresponding author.} \\
Alibaba Cloud Computing \\ \texttt{\{anyue.jgc, liyou.zyw\}@alibaba-inc.com}
}

\begin{document}
\iclrfinalcopy

\maketitle

\begin{abstract}
On-Policy Distillation (OPD) trains a student on its own generations using teacher feedback. With different tokenizers, comparing teacher and student predictions requires alignment at both sequence and vocabulary levels. In this paper, we examine whether expanding this alignment coverage improves learning. Across three heterogeneous teacher--student pairs on mathematical reasoning and code generation, strict $1{:}1$ groups already cover most student-generated tokens despite substantial vocabulary mismatch. On responses sampled from the students before distillation, the shared vocabulary retains nearly all teacher and student probability mass at strictly aligned positions on average. Restricting reverse KL to a student-selected top-16 subset of the shared vocabulary at each strict position achieves accuracy comparable to full shared-vocabulary OPD, outperforming the evaluated cross-tokenizer baselines. Adding mean squared error supervision on span log-probabilities in mismatch groups gives complete supervision coverage, yet reduces accuracy. At checkpoints from training with only the strict loss, the span gradients show weak or negative directional agreement with the strict gradients and grow in magnitude relative to them. These diagnostics may help explain the accuracy drop from adding span supervision. Our findings motivate a shift from maximizing \emph{alignment coverage} to prioritizing \emph{supervision reliability}: compact supervision at strict positions can be more effective than broader coverage that introduces weakly aligned or conflicting training signals.
\end{abstract}

\begin{figure}[!ht]
    \centering
    \includegraphics[width=\linewidth]{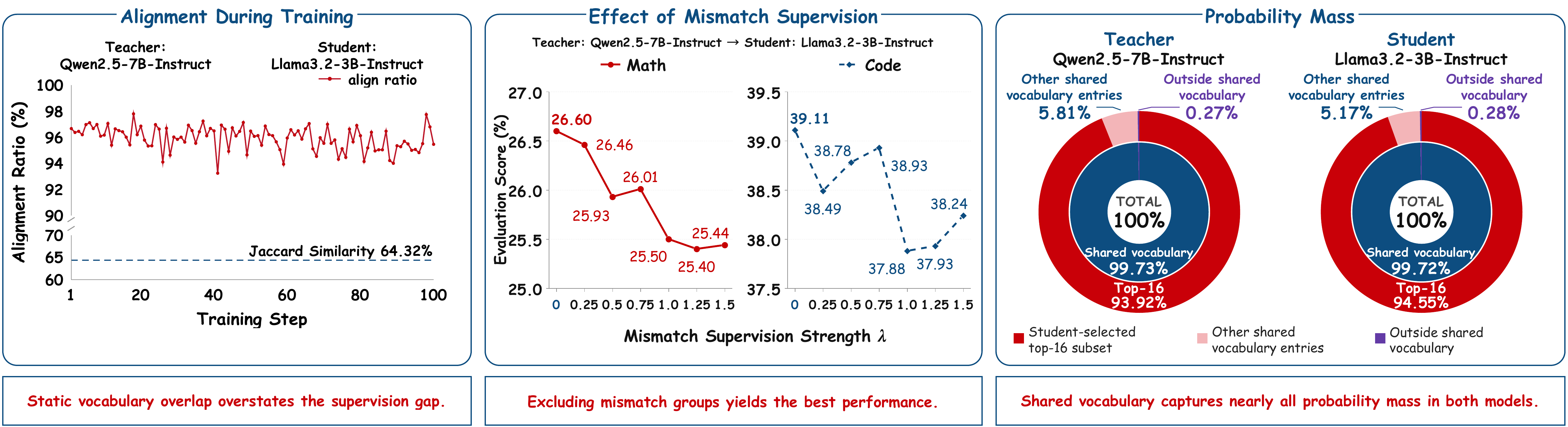}
     \caption{Three diagnostics of Cross-Tokenizer OPD for Qwen2.5-7B-Instruct$\rightarrow$Llama-3.2-3B-Instruct. \textbf{Left:} The logged student-side alignment ratio remains high during Strict full training despite substantially lower static vocabulary overlap. \textbf{Middle:} Adding supervision on mismatch groups lowers mathematics and code performance. \textbf{Right:} At strictly aligned positions in student-generated responses before distillation, the shared vocabulary retains nearly all teacher and student probability mass on average, and a student-selected top-16 subset retains most of it on both sides.}
    \label{fig:coverage-reliability-overview}
\end{figure}

\section{Introduction}
\label{sec:introduction}

On-Policy Distillation (OPD) trains a language model on its own generations using feedback from a teacher \citep{DBLP:conf/iclr/AgarwalVZSGGB24,gu2024minillm}. By supplying supervision at the prefixes reached by the current student, OPD connects teacher feedback to the student's evolving behavior. Extending this method across model families requires comparing predictions from models with different tokenizers. The same response can have different token boundaries, and the next token distributions are defined over different vocabularies. Cross-Tokenizer OPD must therefore account for alignment at both sequence and vocabulary levels.

Recently, Cross-Tokenizer distillation addresses these obstacles through rank-based distribution matching, learned mappings, likelihood matching, and byte-level interfaces \citep{boizard2024uld,zhang2024dualspace,minixhofer2025alm,singh2026bytelevel}. For on-policy training, SimCT recovers supervision through aligned multi-token units and reports gains over shared-vocabulary OPD \citep{sun2026simct}. Byte-Prefix Marginalization maps teacher probability mass onto student tokens, while also identifying harmful supervision at whitespace positions \citep{wang2026cross}. These developments make additional teacher signals accessible and motivate examining their learning value. We ask: \emph{how much useful supervision does strict matching already retain, and what does recovering the excluded targets contribute to learning?}

\emph{Alignment coverage} describes how much of the response and the models' vocabularies can be compared across tokenizers. \emph{Supervision reliability} concerns whether the resulting teacher targets provide useful guidance for student learning at the states it visits. Static vocabulary overlap counts vocabulary entries equally, whereas their frequency and predictive probability on student trajectories can be highly uneven. A large vocabulary gap can therefore coexist with frequent strict alignment and concentrated probability mass on shared vocabulary entries. For spans that lack strict alignment, multi-token grouping makes additional supervision possible. Whether this additional supervision improves learning depends on the objective applied to the aligned groups.

Prior OPD studies show that concentrated prediction supports can retain substantial distillation benefits \citep{fu2026revisiting,li2026rethinking}. Under heterogeneous tokenizers, evaluating such supports also requires accounting for which response positions and vocabulary entries admit direct comparison. We evaluate the learning value of this supervision through training comparisons, using probability-mass measurements and gradient diagnostics to interpret the outcomes.

Our study covers three heterogeneous teacher--student pairs on mathematical reasoning and code generation, including instruction-tuned and base students. We start from strict Cross-Tokenizer OPD, which applies reverse KL at strictly aligned positions. At these positions, one student token and one teacher token cover the same span of the response. Both predictive distributions are restricted and renormalized on the shared vocabulary. We then vary the weight of a mean squared error (MSE) loss on span log-probabilities and the number of shared vocabulary entries used at each strict position. Figure~\ref{fig:coverage-reliability-overview} shows the coverage, mismatch-supervision, and probability-mass observations. Our main findings are:

\begin{itemize}
    \item \textbf{Closing the coverage gap with span MSE reduces accuracy.} Strict student-token coverage is 85.57--96.98\% across the studied runs, despite static vocabulary Jaccard overlap of 39.49--64.87\%. Holding the strict objective fixed, we add MSE supervision on span log-probabilities. Every positive weight gives complete structural supervision coverage, yet all 18 tested positive-weight settings lower the full-average accuracy (Section~\ref{sec:dynamic-coverage}).

    \item \textbf{A student-selected top-$k$ subset retains most of the learning benefit.} On responses sampled from the students before distillation, the shared vocabulary retains nearly all teacher and student probability mass at strict positions on average. The selected subsets also retain most of this mass for both models. With $k=16$, training preserves at least 96\% of the full-average improvement achieved by full shared-vocabulary OPD over the undistilled student. The full and compact strict objectives attain higher full averages than the evaluated cross-tokenizer baselines on all three model pairs (Section~\ref{sec:method}).

    \item \textbf{Span gradients have weak directional agreement and growing relative magnitude.} At checkpoints from training with only the strict loss, the span gradients show weak or negative agreement with strict gradients. Their agreement is consistently below a reference formed by splitting the positions contributing to the strict loss into two subsets, and the span-to-strict gradient norm ratio increases across the measured training checkpoints. The growing relative scale of these weakly aligned gradients may help explain the lower accuracy under span supervision (Section~\ref{sec:gradient-analysis}).
\end{itemize}

A student-selected top-$k$ subset of the shared vocabulary retains most of the observed distillation gains, while the span objective expands structural supervision coverage at the cost of accuracy. These findings motivate a shift from \emph{alignment coverage} to \emph{supervision reliability}: the value of additional supervision should be assessed through its interaction with strict distribution matching and its effect on downstream performance.

\section{Preliminaries}
\label{sec:problem-setup}

\subsection{Notation}

Let $x \sim \mathcal{D}$ denote a \textit{prompt} and $y = (y_1, \ldots, y_L)$ denote a \textit{response} token sequence, where $y_{<i} = (y_1, \ldots, y_{i-1})$ is the \textit{prefix} of $y$. We consider two LLMs: a student model $\pi_\theta$ and a frozen teacher model $\pi_{\text{T}}$, where $\pi_\theta$ defines a distribution $\pi_\theta(\cdot | x)$ over vocabulary $\mathcal{V}_\theta$, and $\pi_\text{T}$ defines a distribution $\pi_\text{T}(\cdot | x)$ over vocabulary $\mathcal{V}_\text{T}$. A \textit{trajectory} is a pair $(x, y)$ with $y$ sampled autoregressively from the policy given $x$.

\subsection{On-Policy Distillation}

On-Policy Distillation (OPD) samples trajectories from the student and aligns the student to the teacher on the prefixes the student actually visits \citep{DBLP:conf/iclr/AgarwalVZSGGB24}. OPD can also be viewed as a special case of dense KL-constrained reinforcement learning, where the teacher distribution induces a token-level reward and the KL regularizer has a fixed relative weight \citep{yang2026learning}. With a shared tokenizer, OPD minimizes the following objective:
\begin{align}
    \mathcal{L}_\text{OPD} (\theta) = \mathbb{E}_{x \sim \mathcal{D}, y \sim \pi_\theta (\cdot | x)} \left[ \sum_{i=1}^L \text{KL}(\pi_\theta (\cdot | x, y_{<i}) \| \pi_\text{T} (\cdot | x, y_{<i})) \right].
\end{align}

\subsection{Cross-Tokenizer On-Policy Distillation}

Different tokenizers can assign different token boundaries and vocabulary entries to the same text. Cross-Tokenizer OPD must therefore account for alignment at both sequence and vocabulary levels \citep{wang2026cross}.

\paragraph{Token-group alignment.} For scoring and alignment, we tokenize the decoded student response with both models' tokenizers. We denote the resulting student and teacher response-token sequences by $y=(y_1,\ldots,y_L)$ and $v=(v_1,\ldots,v_n)$, respectively. The implementation preserves the positions of these response tokens in each model's encoded input, as detailed in Appendix~\ref{app:experimental-details}. We then retain the token offsets shared by both sequences, including the start and end of the response. For the $r$-th interval between consecutive shared boundaries, let $S_r^\theta$ and $S_r^{\text{T}}$ contain the student and teacher token indices within that interval, respectively. The tokens in these two aligned token groups concatenate to the same span. Applying this construction to every interval partitions the response into $R$ pairs of aligned token groups:
\begin{align}
    \mathcal{S}(y) =\left\{\bigl(S_r^\theta,S_r^{\text{T}}\bigr)\right\}_{r=1}^{R}.
    \label{eq:alignment-groups}
\end{align}
We divide the group indices into those of strict $1{:}1$ groups and mismatch groups:
\begin{equation}
    \mathcal{A}_{1{:}1}(y) =\left\{r:|S_r^\theta|=|S_r^{\text{T}}|=1\right\}, \quad
    \mathcal{A}_{\text{mis}}(y) =\{1,\ldots,R\}\setminus\mathcal{A}_{1{:}1}(y).
    \label{eq:strict-groups}
\end{equation}
A strict $1{:}1$ group consists of one student token and one teacher token spanning the same interval of the response. We refer to their paired prediction positions as \emph{strict positions}. A mismatch group requires multiple tokens on at least one side to construct the same span.

\paragraph{Shared vocabulary.} We identify vocabulary entries across models by their underlying tokens, excluding special tokens and ambiguous matches. Under this convention, the shared vocabulary is $\mathcal{V}_{\cap}=\mathcal{V}_\theta\cap\mathcal{V}_{\text{T}}$. For each $r\in\mathcal{A}_{1{:}1}(y)$, write $S_r^\theta=\{i_r\}$ and $S_r^{\text{T}}=\{j_r\}$. The student and teacher predict according to $\pi_\theta(\cdot\mid x,y_{<i_r})$ and $\pi_{\text{T}}(\cdot\mid x,v_{<j_r})$, respectively. For $w\in\mathcal{V}_{\cap}$, we use $w$ to index its corresponding vocabulary entry in each distribution. Restricting and renormalizing both distributions over the shared vocabulary gives
\begin{equation}
    \bar\pi_\theta(w\mid x,y_{<i_r}) = \frac{\pi_\theta(w\mid x,y_{<i_r})}{\sum_{w'\in\mathcal{V}_{\cap}}\pi_\theta(w'\mid x,y_{<i_r})},
    \bar\pi_{\text{T}}(w\mid x,v_{<j_r}) = \frac{\pi_{\text{T}}(w\mid x,v_{<j_r})}{\sum_{w'\in\mathcal{V}_{\cap}}\pi_{\text{T}}(w'\mid x,v_{<j_r})}.
    \label{eq:shared-renormalization}
\end{equation}

\paragraph{Strict Cross-Tokenizer objective.} Following the OPD objective above, we sum reverse KL over strictly aligned positions using the shared-vocabulary distributions:
\begin{equation}
    \mathcal{L}_{1{:}1}(\theta) = \mathbb{E}_{x\sim\mathcal{D},\,y\sim\pi_\theta(\cdot\mid x)}\left[ \sum_{r\in\mathcal{A}_{1{:}1}(y)} \text{KL}\left(\bar\pi_\theta(\cdot\mid x,y_{<i_r}) \| \bar\pi_{\text{T}}(\cdot\mid x,v_{<j_r})\right) \right].
    \label{eq:strict-opd-loss}
\end{equation}

\section{Revisiting Supervision Coverage}
\label{sec:dynamic-coverage}

Figure~\ref{fig:coverage-reliability-overview} presents three observations for Qwen$\rightarrow$Llama: strict alignment remains high during training, adding span supervision lowers the reported downstream scores, and most predictive mass at strict positions lies on shared vocabulary entries. In this section, we first measure strict token coverage, then test the learning effect of supervising mismatch groups, and finally measure the probability mass retained by the shared vocabulary.

\paragraph{Experimental setting.} The experiments cover four model series: Qwen \citep{yang2025qwen3}, Llama \citep{grattafiori2024llama}, Granite\footnote{\url{https://huggingface.co/blog/ibm-granite/granite-4-1}}, and Phi \citep{abouelenin2025phi}. Throughout, arrows point from teacher to student. We study Qwen2.5-7B-Instruct$\rightarrow$Llama-3.2-3B-Instruct, Granite-4.1-8B$\rightarrow$Phi-4-mini-instruct, and Granite-4.1-8B$\rightarrow$Qwen2.5-7B-Base, abbreviated as Qwen$\rightarrow$Llama, Granite$\rightarrow$Phi, and Granite$\rightarrow$Qwen, respectively. Training starts from a common source pool of 20,000 prompts, comprising 10,000 mathematics prompts from DAPO-Math-17K \citep{dapo} and 10,000 code prompts from CodeForces\footnote{\url{https://huggingface.co/datasets/open-r1/codeforces}}. We evaluate mathematics on MATH500 \citep{math500}, GSM8K \citep{gsm8k}, AIME-2024, AIME-2025, AIME-2026, AMC23, and Minerva-Math \citep{minerva}, and code on HumanEval \citep{humaneval}, MBPP \citep{mbpp}, and LiveCodeBench \citep{livecodebench}. We report accuracy on all benchmarks, averaging over 32 sampled responses per problem for mathematics (mean@32) and 8 for code (mean@8). The \emph{full} gives equal weight to the arithmetic means within the two domains. Appendix~\ref{app:experimental-details} provides experimental details.

\subsection{Strict Alignment on Student Trajectories}
\label{sec:trajectory-coverage}

\paragraph{Static overlap and dynamic coverage.} We compare static vocabulary Jaccard overlap $J_{\text{vocab}}$ with strict token coverage on student rollouts. Appendix~\ref{app:detailed-diagnostics} gives the static overlap definition and complete vocabulary statistics. For the response $y$, strict student-token coverage is the fraction of non-special student tokens belonging to strict $1{:}1$ groups:
\begin{equation}
    C_{1{:}1}^{\theta}(y)
    = \frac{\sum_{r\in\mathcal{A}_{1{:}1}(y)} |S_r^\theta|}
    {\sum_{r=1}^{R} |S_r^\theta|}.
    \label{eq:strict-coverage}
\end{equation}
Strict teacher-token coverage $C_{1{:}1}^{\text{T}}(y)$ replaces $S_r^\theta$ with $S_r^{\text{T}}$ in both sums.

\begin{table}[t]
    \caption{Static vocabulary Jaccard overlap $J_{\text{vocab}}$ and strict token coverage on student trajectories from Strict full training. $C_{1{:}1}^{\theta}$ and $C_{1{:}1}^{\mathrm{T}}$ denote student and teacher strict token coverage over the indicated training steps.}
    \label{tab:dynamic-coverage}
    \begin{center}
    \begin{tabular}{lrlrrrr}
        \toprule
        \multirow{2}{*}{\textbf{Teacher}$\rightarrow$\textbf{Student}}
        & \multirow{2}{*}{$J_{\text{vocab}}$}
        & \multirow{2}{*}{\textbf{Coverage}}
        & \multicolumn{4}{c}{\textbf{Training steps}} \\
        \cmidrule(lr){4-7}
        & & & \textbf{1--100} & \textbf{1--20} & \textbf{41--60} & \textbf{81--100} \\
        \midrule
        \multirow{2}{*}{Qwen$\rightarrow$Llama}
        & \multirow{2}{*}{64.32}
        & $C_{1{:}1}^{\theta}$ & 93.56 & 94.63 & 93.20 & 93.43 \\
        & & $C_{1{:}1}^{\mathrm{T}}$ & 85.91 & 88.65 & 84.90 & 85.36 \\
        \addlinespace
        \multirow{2}{*}{Granite$\rightarrow$Phi}
        & \multirow{2}{*}{39.49}
        & $C_{1{:}1}^{\theta}$ & 96.98 & 96.65 & 96.95 & 97.12 \\
        & & $C_{1{:}1}^{\mathrm{T}}$ & 97.26 & 95.83 & 97.50 & 97.96 \\
        \addlinespace
        \multirow{2}{*}{Granite$\rightarrow$Qwen}
        & \multirow{2}{*}{64.87}
        & $C_{1{:}1}^{\theta}$ & 85.57 & 85.23 & 85.06 & 85.88 \\
        & & $C_{1{:}1}^{\mathrm{T}}$ & 82.82 & 81.15 & 82.77 & 83.86 \\
        \bottomrule
    \end{tabular}
    \end{center}
\end{table}

\paragraph{Strict supervision remains prevalent.} All 512 student rollouts generated at each of the 100 iterations are saved, giving 51,200 responses per run. We aggregate coverage over the complete run and three disjoint windows: steps 1--20, 41--60, and 81--100. Table~\ref{tab:dynamic-coverage} contrasts Jaccard overlap of 39.49--64.87\% with strict token coverage of 85.57--96.98\% of student tokens and 82.82--97.26\% of teacher tokens over the complete runs. Granite$\rightarrow$Phi has the lowest Jaccard overlap but the highest coverage under both tokenizations. Within each pair, the range across the three reported windows is at most 1.43 percentage points for strict student-token coverage and 3.75 percentage points for strict teacher-token coverage. These results show that substantial static vocabulary mismatch can coexist with strict alignment at most positions on student-generated trajectories. High strict token coverage alone, however, does not establish whether the remaining mismatch groups offer useful supervision.

\subsection{Learning Value of Mismatch Supervision}
\label{sec:mismatch-performance}
\begin{figure}[!t]
    \centering
    \includegraphics[width=\linewidth]{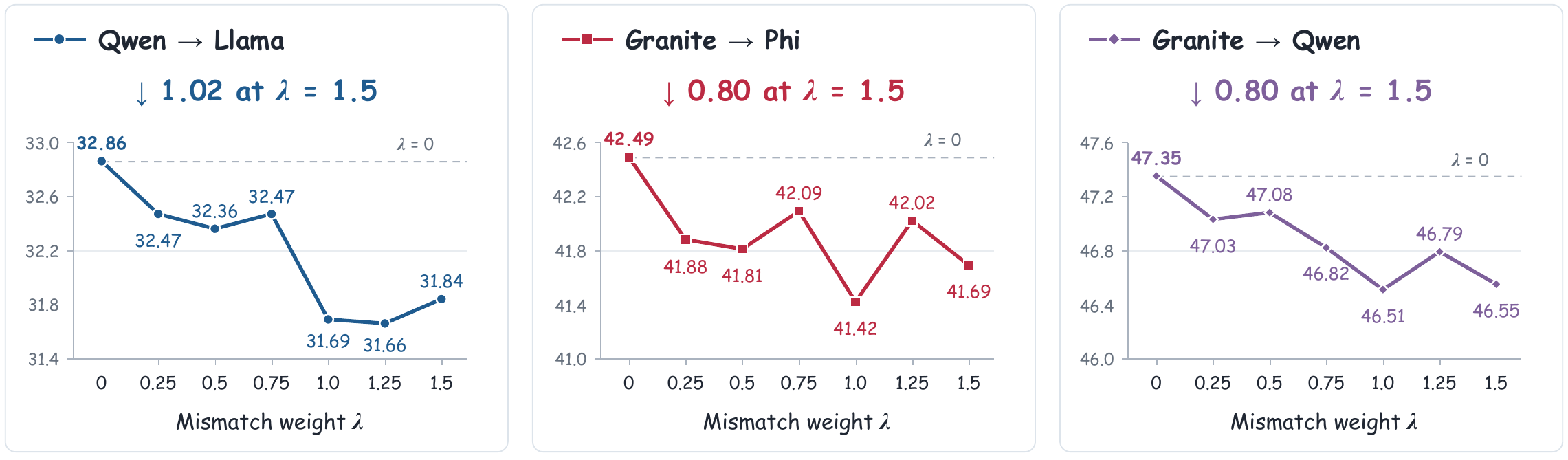}
    \caption{Full-average accuracy (\%) across mismatch weights. Dashed lines mark strict supervision ($\lambda=0$), and annotations give the decrease at $\lambda=1.5$ in percentage points.}
    \label{fig:mismatch_weight}
\end{figure}

We test the learning value of the remaining mismatch groups by adding span supervision to the strict objective in \Eqref{eq:strict-opd-loss}.

\paragraph{Span supervision.} Each mismatch group in \Eqref{eq:alignment-groups} pairs student and teacher token sequences for the same text span, potentially involving different numbers of autoregressive decisions. For $r\in\mathcal{A}_{\text{mis}}(y)$, we compute the probabilities of these observed token paths as
\begin{equation}
    q_\theta^{(r)} = \prod_{i\in S_r^\theta}\pi_\theta(y_i\mid x,y_{<i}), \quad q_{\text{T}}^{(r)} = \prod_{j\in S_r^{\text{T}}}\pi_{\text{T}}(v_j\mid x,v_{<j}).
    \label{eq:observed-span-probability}
\end{equation}
We match these probabilities with a mean squared error (MSE) loss over mismatch groups, using a log-probability formulation:
\begin{equation}
    \mathcal{L}_\text{span}(\theta) = \mathbb{E}_{x \sim \mathcal{D}, y \sim \pi_\theta(\cdot | x)} \left[ \sum_{r \in \mathcal{A}_\text{mis}(y)} \left( \log q_\theta^{(r)} - \log q_\text{T}^{(r)} \right)^2 \right].
    \label{eq:span-mse-loss}
\end{equation}
The total loss is
\begin{equation}
    \mathcal{L}_\lambda (\theta) = \mathcal{L}_{1{:}1}(\theta) + \lambda \mathcal{L}_\text{span}(\theta).
    \label{eq:expanded-opd-loss}
\end{equation}
The baseline $\lambda=0$ applies supervision only to strict $1{:}1$ groups. Every $\lambda>0$ additionally supervises all mismatch groups. All positive weights therefore give 100\% complete supervision coverage. Varying $\lambda$ changes the influence of the span loss while keeping the set of included groups fixed for any given response. We sweep $\lambda\in\{0,0.25,0.5,0.75,1.0,1.25,1.5\}$ for all three teacher--student pairs, as shown in Figure~\ref{fig:mismatch_weight}.

% % Required packages in the preamble:
% % \usepackage{booktabs}
% % \usepackage[table]{xcolor}

% \begin{table}[t]
%     \caption{Full-average accuracy (\%) with MSE supervision on observed-span log probabilities. The metric gives equal weight to math and code averages. All positive weights include every mismatch group. Bold marks the highest score within each row.}
%     \label{tab:mismatch-ablation-summary}

%     \definecolor{TableHeader}{HTML}{E7EDF4}
%     \definecolor{TableSubheader}{HTML}{F3F6FA}
%     \definecolor{TableBest}{HTML}{D3E9DD}

%     \begin{center}
%     \begin{tabular}{l rrrrrrr}
%         \toprule

%         \rowcolor{TableHeader}
%         \textbf{Model pair}
%         & \multicolumn{7}{c}{
%             \cellcolor{TableHeader}
%             \textbf{Mismatch weight $\lambda$}
%         } \\

%         \cmidrule(lr){2-8}

%         \rowcolor{TableSubheader}
%         & $0$ & $0.25$ & $0.5$ & $0.75$ & $1.0$ & $1.25$ & $1.5$ \\

%         \midrule

%         Qwen$\rightarrow$Llama
%         & \cellcolor{TableBest}\textbf{32.86}
%         & 32.47 & 32.36 & 32.47 & 31.69 & 31.66 & 31.84 \\

%         Granite$\rightarrow$Phi
%         & \cellcolor{TableBest}\textbf{42.49}
%         & 41.88
%         & 41.81
%         & 42.09
%         & 41.42
%         & 42.02
%         & 41.69 \\

%         Granite$\rightarrow$Qwen
%         & \cellcolor{TableBest}\textbf{47.35}
%         & 47.03
%         & 47.08
%         & 46.82
%         & 46.51
%         & 46.79
%         & 46.55 \\

%         \bottomrule
%     \end{tabular}
%     \end{center}
% \end{table}

\paragraph{Strict supervision performs best throughout the weight sweep.} All three pairs attain their highest full average at $\lambda=0$. The 18 positive-weight settings score 0.27--1.20 percentage points below their respective strict baselines. The detailed results in Appendix~\ref{app:complete-results} show that $\lambda=0$ also gives the highest math and code averages for each pair. Figure~\ref{fig:mismatch_weight} shows an overall downward trend in full-average accuracy as $\lambda$ increases across all three pairs. With span log-probability MSE, achieving complete supervision coverage reduces downstream accuracy across the tested weights. This extends the distinction between coverage and learning value in Figure~\ref{fig:coverage-reliability-overview} to all three model pairs.

\subsection{Predictive Mass Concentrates on the Shared Vocabulary}
\label{sec:shared-support-mass}

We now measure the probability mass that each model assigns to the shared vocabulary at strict positions. Static vocabulary overlap counts vocabulary entries, but their contribution to the predictive distributions depends on the probability assigned to them in context.

\paragraph{Measuring probability mass.} For each strict $1{:}1$ group $r\in\mathcal{A}_{1{:}1}(y)$, abbreviate the original full-vocabulary predictions at its strict positions as $\pi_\theta^{(r)}(w)=\pi_\theta(w\mid x,y_{<i_r})$ and $\pi_{\text{T}}^{(r)}(w)=\pi_{\text{T}}(w\mid x,v_{<j_r})$. For $a\in\{\theta,\text{T}\}$, the probability mass on the shared vocabulary is
\begin{equation}
    M_{\cap,a}^{(r)} = \sum_{w\in\mathcal{V}_{\cap}}\pi_a^{(r)}(w).
    \label{eq:shared-vocab-mass}
\end{equation}
To measure concentration within the shared vocabulary, we select top-$k$ shared vocabulary entries to which it assigns the highest probabilities. These entries form the \emph{student-selected top-$k$ subset}:
\begin{equation}
    \mathcal{K}_{k}^{(r)} = \text{TopK}_{w\in\mathcal{V}_{\cap}} \left(\pi_\theta^{(r)}(w), k\right), \quad M_{k,a}^{(r)} = \sum_{w\in\mathcal{K}_{k}^{(r)}}\pi_a^{(r)}(w).
    \label{eq:student-selected-mass}
\end{equation}
Unlike the shared vocabulary, this subset depends on the student distribution at each strict position. When used as the support of the distillation loss, we also refer to it as the \emph{top-$k$ support}.

\begin{wrapfigure}{r}{0.5\textwidth}
    \centering
    \includegraphics[width=\linewidth]{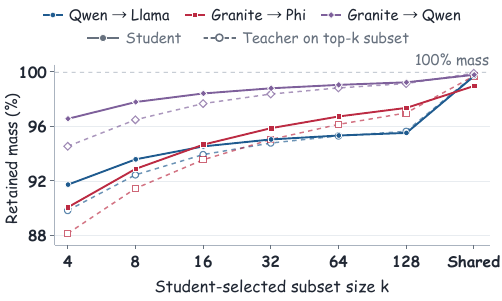}
    \caption{Probability mass retained by student-selected top-$k$ subsets and the full shared vocabulary at strict positions before distillation.}
    \label{fig:compact-support-mass}
\end{wrapfigure}

For each model pair, the probability-mass probe uses all 20k prompts. We use the student before distillation to generate one rollout per prompt, scoring both models at each measured strict position. Both models are evaluated on the same student-selected top-$k$ subset using their original softmax probabilities. We average each statistic over all measured strict positions and evaluate $k\in\{4,8,16,32,64,128\}$ alongside the full shared vocabulary. Appendix~\ref{app:detailed-diagnostics} reports the detailed results.

\paragraph{Shared vocabulary captures nearly all predictive mass.} The ``Shared'' endpoints in Figure~\ref{fig:compact-support-mass} show that the full shared vocabulary retains 99.69--99.90\% of teacher mass and 98.99--99.81\% of student mass on average across the three pairs. Strict alignment makes supervision available at most generated token positions, while shared vocabulary entries carry almost all predictive probability at the measured strict positions. The vocabulary entries excluded from comparison collectively receive little probability on average under either model. Therefore, static vocabulary mismatch alone does not determine how much probability mass the shared vocabulary retains.

\paragraph{A student-selected top-$k$ subset also captures most teacher mass.} At $k=16$, the selected shared vocabulary entries retain at least 93.54\% of teacher mass and 94.55\% of student mass on average in each of the three pairs. From $k=4$ to $k=128$, successive doublings of the subset size yield progressively smaller gains in all six curves. Expanding the subset from 16 to 128 entries increases the retained mass by at most 3.44 percentage points for the teacher and 2.70 points for the student. Selection uses only the student distribution, yet the selected shared vocabulary entries also carry most of the teacher's predictive mass.

\section{Learning with Student-Selected Subsets}
\label{sec:method}

In this section, we test whether training on this subset retains the learning gains of distillation on the full shared vocabulary.

We restrict the strict objective in \Eqref{eq:strict-opd-loss} to the student-selected top-$k$ subset $\mathcal{K}_k^{(r)}$, renormalizing both models' distributions on this same set before computing reverse KL. We compare $k=16$ and $k=128$ with the full shared vocabulary across all three model pairs and report evaluation results on mathematical reasoning and code generation benchmarks. We also compare strict supervision with ULD \citep{boizard2024uld}, its span-grouped variant Extended ULD, GOLD\footnote{\url{https://huggingface.co/docs/trl/gold_trainer}}, and SimCT \citep{sun2026simct}. Table~\ref{tab:main-results} reports domain and full averages for all methods, and Appendix~\ref{app:complete-results} provides benchmark-level results. All four baselines are implemented in KDFlow \citep{zhang2026kdflow}. Appendix~\ref{app:experimental-details} details their position alignment, vocabulary handling, EOS settings, and loss reductions.

\begin{table}[t]
    \caption{Cross-tokenizer distillation accuracy (\%). Base denotes the student before distillation. Bold marks the best distilled score per column.}
    \label{tab:main-results}

    \begin{center}
    \begin{tabular}{lrrrrrrrrr}
        \toprule
        \multirow{2}{*}{\textbf{Method}} & \multicolumn{3}{c}{\textbf{Qwen$\rightarrow$Llama}} & \multicolumn{3}{c}{\textbf{Granite$\rightarrow$Phi}} & \multicolumn{3}{c}{\textbf{Granite$\rightarrow$Qwen}} \\
        \cmidrule(lr){2-4}\cmidrule(lr){5-7}\cmidrule(lr){8-10}
        & \textbf{Math} & \textbf{Code} & \textbf{Full} & \textbf{Math} & \textbf{Code} & \textbf{Full} & \textbf{Math} & \textbf{Code} & \textbf{Full} \\
        \midrule
        Base & 18.32 & 35.60 & 26.96 & 33.06 & 40.04 & 36.55 & 24.61 & 34.51 & 29.56 \\
        \midrule
        ULD & 20.51 & 37.74 & 29.12 & 35.24 & 42.01 & 38.63 & 33.36 & 20.87 & 27.12 \\
        Extended ULD & 23.01 & \textbf{39.26} & 31.13 & 35.28 & 41.42 & 38.35 & 33.14 & 32.00 & 32.57 \\
        GOLD & 21.21 & 33.10 & 27.16 & 36.22 & 47.29 & 41.75 & 38.26 & 54.84 & 46.55 \\
        SimCT & 25.31 & 37.86 & 31.59 & 36.67 & 46.70 & 41.68 & 38.64 & 53.56 & 46.10 \\
        \midrule
        Strict full & 26.60 & 39.11 & \textbf{32.86} & \textbf{37.49} & 47.49 & 42.49 & \textbf{39.85} & 54.84 & 47.35 \\
        Strict top-16 & \textbf{26.86} & 38.42 & 32.64 & 37.23 & 47.29 & 42.26 & 39.07 & 55.06 & 47.06 \\
        Strict top-128 & 26.31 & 38.44 & 32.38 & 37.27 & \textbf{47.82} & \textbf{42.54} & 39.67 & \textbf{55.62} & \textbf{47.65} \\
        \bottomrule
    \end{tabular}
    \end{center}
\end{table}

\paragraph{Strict supervision improves both domains.} Strict full and both compact variants improve math and code accuracy over Base in every model pair, providing the reference for assessing how much learning benefit is retained on the student-selected top-$k$ subsets.

\paragraph{Top-$k$ subsets retain most of the performance gains.} Top-16 retains at least 96\% of the reported full-average improvement from Base to strict full in each pair. Most of the observed learning benefit can therefore be retained on a small student-selected top-$k$ subset, with no consistent further gain from increasing its size.

\paragraph{Compact supervision achieves higher performance.} All three strict variants exceed all four alternatives in both math and full averages on every model pair, with top-16 ahead of the strongest alternative in the full average by 0.51--1.05 percentage points. Appendix~\ref{app:qwen3-granite} reports additional results for distillation from a larger 235B teacher to an 8B student, where Strict top-16 achieves the highest Math, Code, and Full averages among the evaluated distillation methods. On ALFWorld, Strict top-16 also attains the highest success rate among the evaluated distillation methods on both seen and unseen splits, as shown in Appendix~\ref{app:alfworld}.

\section{Optimization Diagnostics of Span Supervision}
\label{sec:gradient-analysis}

In this section, we examine the gradient direction and relative scale between $\mathcal{L}_{1:1}(\theta)$ and $\mathcal{L}_\text{span}(\theta)$ in \Eqref{eq:expanded-opd-loss}.

\paragraph{Gradient diagnostics.} We compare the gradients of strict distribution matching and the span loss, evaluating both losses on the same sampled responses held fixed during differentiation:
\begin{equation}
    g_{1{:}1} = \nabla_\theta \mathcal{L}_{1{:}1}, \quad g_{\text{mis}} = \nabla_\theta\mathcal{L}_{\text{span}}.
    \label{eq:gradient-components}
\end{equation}
We measure directional agreement using gradient cosine \citep{yu2020gradient} and relative magnitude using the norm ratio:
\begin{equation}
    c_{\text{mis}} = \frac{\langle g_{1{:}1},g_{\text{mis}}\rangle}{\lVert g_{1{:}1}\rVert_2\lVert g_{\text{mis}}\rVert_2}, \quad \rho = \frac{\lVert g_{\text{mis}}\rVert_2}{\lVert g_{1{:}1}\rVert_2}.
    \label{eq:gradient-geometry}
\end{equation}
We analyze checkpoints at steps 0, 20, 60, and 100 from the Strict full run for each model pair, generating student responses to the same 512 prompts at each checkpoint. We compute the diagnostics from the raw accumulated gradients over all trainable student parameters, and report the median of each statistic across replay units at each checkpoint. Within each batch, we randomly split the positions contributing to the strict loss into two subsets, $A$ and $B$, and accumulate their gradients separately across the replay unit. Their cosine $c_{\text{ctrl}}=\cos(g_A,g_B)$ provides a reference for gradient agreement between position subsets under the same objective. Table~\ref{tab:gradient-geometry} in Appendix~\ref{app:detailed-diagnostics} reports the complete statistics and replay setting.

\begin{figure}[t]
    \centering
    \begin{minipage}[t]{0.485\linewidth}
        \centering
        \includegraphics[width=\linewidth]{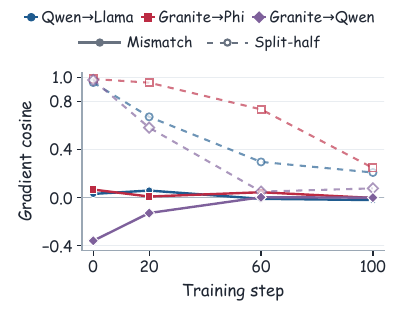}
        \caption{Mismatch--strict gradient cosine $c_{\text{mis}}$ (solid) and split-half reference $c_{\text{ctrl}}$ (dashed) at checkpoints from training with only the strict loss.}
        \label{fig:gradient_direction}
    \end{minipage}\hfill
    \begin{minipage}[t]{0.485\linewidth}
        \centering
        \includegraphics[width=\linewidth]{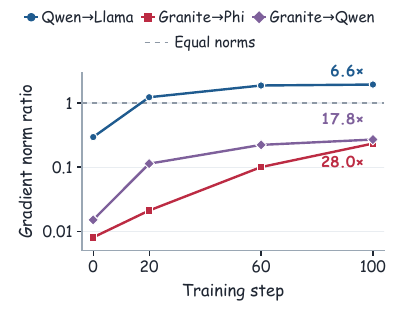}
        \caption{Mismatch-to-strict gradient norm ratio $\rho$, at checkpoints from training with only the strict loss, with growth factors from step 0 to 100 computed from unrounded checkpoint medians.}
        \label{fig:relative_influence}
    \end{minipage}
\end{figure}

\paragraph{Span gradients show weak agreement with strict supervision.} Figure~\ref{fig:gradient_direction} shows mismatch cosines near zero for Qwen$\rightarrow$Llama and Granite$\rightarrow$Phi. Granite$\rightarrow$Qwen instead starts with negative agreement and approaches zero later in training. Although the split-half strict cosine decreases overall during training, it remains higher than the mismatch--strict cosine at every measured checkpoint in each pair.

\paragraph{Relative gradient scale changes during training.} Figure~\ref{fig:relative_influence} shows an increasing span-to-strict gradient norm ratio across the measured checkpoints in all three pairs. At these checkpoints, a fixed positive weight on the span loss would give it an increasing gradient norm relative to the strict component. The growing relative magnitude of these weakly aligned span gradients may contribute to the lower accuracy observed with supervision on mismatch groups in Section~\ref{sec:mismatch-performance}.

\section{Related Work}

\paragraph{On-Policy Distillation.}
On-Policy Distillation trains the student on its own rollouts with teacher supervision \citep{song2026survey}. MiniLLM \citep{gu2024minillm} uses reverse KL to discourage the student from assigning excessive probability to regions unlikely under the teacher. GKD \citep{DBLP:conf/iclr/AgarwalVZSGGB24} supports different divergences and mixtures of student-generated sequences and fixed training data. \citet{yang2026learning} cast OPD as dense KL-constrained RL with an implicit per-token reward given by the teacher-to-reference log-probability ratio. Subsequent work examines which teacher signals are effective at student-visited prefixes. \citet{fu2026revisiting} analyze unreliable guidance and uneven supervision in sampled-token OPD, and propose distribution matching on teacher-selected local top-$k$ supports. Entropy-aware OPD \citep{jin2026entropy} augments reverse KL with forward KL at positions of high teacher entropy, while TIP \citep{xu2026tip} studies trajectory-position selection using student entropy and teacher--student disagreement. \citet{li2026rethinking} relate distillation success to teacher--student compatibility and show that the intersection of their top-$k$ predictions contains most probability mass and can retain the benefits of student top-$k$ supervision. Their shared support consists of high-probability predictions within a common vocabulary. Our comparison of student-selected top-$k$ subsets tests whether concentrated supervision remains sufficient when heterogeneous tokenizers constrain which response positions and vocabulary entries admit direct comparison, using the shared vocabulary at strict positions.

\paragraph{Cross-Tokenizer Distillation.}
Different tokenizers introduce discrepancies in both token boundaries and vocabularies. ULD \citep{boizard2024uld} compares sorted output probabilities without requiring matching token identities. DSKD \citep{zhang2024dualspace} and CDM \citep{chen2025cdm} construct mappings between model representations or vocabularies, while DWA-KD \citep{vu2026dwakd} combines representation alignment with entropy-based weighting. Other methods match likelihoods across tokenizations \citep{minixhofer2025alm} or aggregate tokens into span representations before distillation \citep{dao2026sra}. Byte-Level Distillation converts teacher predictions to byte probabilities and equips the student with a byte-level decoder head \citep{singh2026bytelevel}. For on-policy training, GOLD\footnote{\url{https://huggingface.co/docs/trl/gold_trainer}} merges text-equivalent token groups and uses rank-based or hybrid identity-based distribution matching. SimCT \citep{sun2026simct} adds minimal aligned multi-token units to the shared supervision space and reports improvements over shared-vocabulary OPD from recovering excluded supervision. Byte-Prefix Marginalization \citep{wang2026cross} maps teacher probability mass onto student tokens through byte-prefix relations and retains unmatched mass in a residual category. It also identifies harmful supervision at all-whitespace positions and prevents the resulting training collapse by masking them. We test the learning value of structural recovery by adding an span log-probability MSE loss while holding strict supervision fixed. Probability-mass and gradient analyses, together with comparisons against alternative objectives, distinguish alignment coverage from supervision reliability in the objectives and transfers we evaluate.

\section{Conclusion}
\label{sec:conclusion}

We re-examined the role of alignment coverage in Cross-Tokenizer On-Policy Distillation across three heterogeneous teacher--student pairs on mathematical reasoning and code generation. Strict $1{:}1$ groups already cover most student-generated tokens despite substantial static vocabulary mismatch. On responses sampled from the students before distillation, the shared vocabulary retains nearly all teacher and student probability mass at strict positions on average. A compact student-selected top-$k$ subset of the shared vocabulary retains most of this mass on both sides. Restricting reverse KL to a student-selected top-16 subset of the shared vocabulary at each strict position retains at least 96\% of the full-average improvement achieved by full shared-vocabulary OPD over the undistilled student, with higher performance than the evaluated cross-tokenizer baselines.

Adding span log-probability MSE on mismatch groups achieves complete supervision coverage but reduces downstream accuracy across all tested positive weights. Across the measured checkpoints from training with only the strict loss, the span gradients grow in magnitude relative to the strict gradients, while their directional agreement remains weak or negative. This combination may help explain the accuracy drop from adding span supervision. These findings motivate a shift from \emph{alignment coverage} to \emph{supervision reliability}: increasing coverage alone does not guarantee better distillation, and the quality of the added supervision must be assessed by its contribution to student learning.

\subsection*{AI use statement}

We use generative AI tools solely for language polishing to improve the grammar, clarity, and readability of this paper. These tools were not used for experimental design, implementation, execution, or analysis. We take full responsibility for the final content of this work.

\subsection*{Ethics statement}

\paragraph*{Use of Human Annotations}
Human annotations are only used in methodological research at the beginning of the work, to assist in analyzing the feasibility of the proposed solution. Annotators consented to the use of data for research purposes. We ensure that the privacy of all annotators is protected throughout the annotation process, and all of them are adequately paid according to local standards. Human annotations are not applied during the evaluation of our method.
\paragraph*{Risks}
In this paper, all datasets are obtained from official sources. The datasets adopted have been anonymized and do not contain offensive information. However, we cannot guarantee that the datasets do not contain socially harmful or toxic language.

\subsection*{Reproducibility statement}

We describe the alignment rules, distillation objectives, and diagnostic measures in Sections~\ref{sec:problem-setup}--\ref{sec:gradient-analysis}. Appendix~\ref{app:experimental-details} details the data preparation, training configurations, baseline implementations, and evaluation procedures. Appendix~\ref{app:detailed-diagnostics} provides the protocols for probability-mass measurements and gradient diagnostics. Appendices~\ref{app:complete-results}--\ref{app:alfworld} report the benchmark-level results and additional experiments with a larger teacher and on ALFWorld. Our code is available at \url{https://anonymous.4open.science/r/Cross-Tokenizer-OPD}.

\bibliography{iclr2027_conference}
\bibliographystyle{iclr2027_conference}

\clearpage
\appendix

\section{Experimental Details}
\label{app:experimental-details}

We construct a common source pool of 20,000 prompts for the three teacher--student pairs in Section~\ref{sec:dynamic-coverage}: 10,000 mathematics prompts randomly sampled from DAPO-Math-17K \citep{dapo} and 10,000 code prompts randomly sampled from CodeForces\footnote{\url{https://huggingface.co/datasets/open-r1/codeforces}}. We perform no deduplication. For each student, we encode the formatted student prompt with its tokenizer and retain samples with at most 2,048 prompt tokens. The retained sample count therefore varies with the student tokenizer. The probability-mass probe uses the complete, unfiltered 20k prompts, as detailed in Appendix~\ref{app:detailed-diagnostics}. We use the officially released open-source models named in Section~\ref{sec:dynamic-coverage}. All training prompts are formatted using each model's official chat template.

The current student generates one response per sampled prompt. For loss computation, the student and frozen teacher each re-tokenize the decoded response together with their own prompt encoding. On each side, the response length is measured by separately tokenizing the response without added special tokens, and the prompt-response boundary is then set to the joint input-token count minus that response length. The next-token loss mask selects the resulting response positions. A terminal EOS is appended for responses that were not length-truncated. For the strict objectives, content-token alignment follows Section~\ref{sec:problem-setup}, and paired EOS positions are included in the training loss.

Training runs for 100 on-policy iterations, with checkpoints saved every 10 iterations. Student optimization uses FSDP2 on a single node with 8 NVIDIA H20 GPUs. Sleep mode multiplexes these components on the same workers. We use packed samples, bfloat16 computation, gradient checkpointing, and token-chunked loss evaluation with a chunk size of 2,048 to reduce memory consumption. We use SGLang \citep{sglang} as the rollout engine. The global optimizer batch contains 512 responses across student workers. Table~\ref{tab:training-configuration} lists the optimization, rollout, and training settings.

\begin{table}[ht]
    \caption{Training settings.}
    \label{tab:training-configuration}

    \begin{center}
    \begin{tabular}{ll}
        \toprule
        \textbf{Setting} & \textbf{Value} \\
        \midrule
        \multicolumn{2}{l}{\textbf{Training and optimization}} \\
        Training iterations & 100 \\
        Batch size & 512 \\
        Dynamic batching & Disabled \\
        Optimizer & AdamW \citep{adamw} \\
        Learning rate & $1\times10^{-6}$ \\
        Adam $\beta_1,\beta_2$ & $0.9,0.98$ \\
        Weight decay / grad clip & $0/1.0$ \\
        Checkpoint interval & 10 steps \\
        \midrule
        \multicolumn{2}{l}{\textbf{Distillation and rollout}} \\
        Strict divergence & Reverse KL \\
        KD weight / temperature & $1.0/1.0$ \\
        Responses per prompt & 1 \\
        Rollout sampling & $T=1.0$, top-$p=0.95$ \\
        Max training prompt length & 2,048 \\
        Max generation length & 8,192 \\
        \bottomrule
    \end{tabular}
    \end{center}
\end{table}

\paragraph{Baseline implementations.} All four baselines use the KDFlow \citep{zhang2026kdflow} implementations. SimCT selects \texttt{simct}. All four use unit KD weight without cross-entropy mixing and teacher/student temperature 1.0. ULD, Extended ULD, and GOLD exclude terminal EOS from their distillation losses.

\paragraph{Evaluation.} Evaluation uses publicly available benchmarks. All distilled models in the main comparison and span-MSE sweeps are evaluated at step 100. Mathematics evaluation covers MATH500, GSM8K, AIME-2024, AIME-2025, AIME-2026, AMC23, and Minerva-Math. Code evaluation covers HumanEval, MBPP, and LiveCodeBench. We average correctness over 32 responses per mathematics problem and 8 per code problem, reporting mean@32 and mean@8 accuracy. Math and Code are the arithmetic means over their seven and three benchmarks, respectively, and Full is the equally weighted mean of Math and Code. Appendix~\ref{app:complete-results} reports benchmark-level scores, including the base model and frozen teacher as references. The evaluation clients submit tokenized prompts to SGLang through its OpenAI-compatible completions interface. Both domains use temperature 1.0, top-$p=0.95$, a maximum of 8,192 generated tokens. For mathematics, the input is the benchmark file's \texttt{problem} field as a single user message, rendered with the model's chat template and generation prompt. For HumanEval and MBPP, a single user message contains the dataset prompt followed by an instruction to think through the problem and place the Python solution in a final fenced code block. For LiveCodeBench, we use its generic system message and question template. Each prompt is encoded without adding further special tokens after template rendering.

\paragraph{Mathematics scoring.} The grader extracts the final answer from each response. A missing or malformed final boxed answer is counted as incorrect. If the extracted answer exceeds 300 characters, only its first 300 characters are passed to the grader. Both the reference and prediction are wrapped as boxed expressions and processed using Math-Verify\footnote{\url{https://github.com/huggingface/Math-Verify}}. Parsing or verification exceptions count as incorrect. Failed generation requests yield empty responses and remain in the accuracy denominator.

\paragraph{Code scoring.} HumanEval and MBPP are loaded through EvalPlus \citep{evoplus}. The evaluator computes both original and augmented-test outcomes. The reported HumanEval and MBPP scores use the original base tests. For extraction, it takes the text between the final two code-fence lines, or at most the first 200 lines when no complete fence is found, then applies EvalPlus sanitization with the task entry point. LiveCodeBench uses \texttt{release\_v6} version, with 1,055 problems, code extractor, and code-generation test runner. The wrapper supplies an execution timeout of 10 seconds to the runner and additionally caps each candidate's total evaluation wait at 120 seconds. Scores are averaged from each problem's fraction of passing candidates.

\section{Diagnostic Protocols and Statistics}
\label{app:detailed-diagnostics}

Using the vocabulary convention in Section~\ref{sec:problem-setup}, we measure static vocabulary overlap by
\begin{equation}
    J_{\text{vocab}} = \frac{|\mathcal{V}_{\cap}|}{|\mathcal{V}_{\theta}\cup\mathcal{V}_{\mathrm{T}}|} = \frac{|\mathcal{V}_\theta \cap \mathcal{V}_\mathrm{T}|}{|\mathcal{V}_{\theta}\cup\mathcal{V}_{\mathrm{T}}|}.
    \label{eq:vocab-jaccard}
\end{equation}
Table~\ref{tab:static-vocab-overlap} reports the vocabulary sizes and the fractions of the student and teacher vocabularies in the shared vocabulary.

\begin{table}[t]
    \caption{Static vocabulary overlap across the three teacher--student pairs. Vocabulary counts follow the shared-vocabulary convention in Section~\ref{sec:problem-setup}. Overlap ratios are reported as percentages.}
    \label{tab:static-vocab-overlap}
    \begin{center}
    \begin{tabular}{lrrrrrr}
        \toprule
        \multirow{2}{*}{\textbf{Teacher}$\rightarrow$\textbf{Student}}
        & \multicolumn{3}{c}{\textbf{Vocabulary size}}
        & \multicolumn{3}{c}{\textbf{Overlap (\%)}} \\
        \cmidrule(lr){2-4}
        \cmidrule(lr){5-7}
        & $|\mathcal{V}_{\mathrm{T}}|$
        & $|\mathcal{V}_{\theta}|$
        & $|\mathcal{V}_{\cap}|$
        & $\frac{|\mathcal{V}_{\cap}|}{|\mathcal{V}_{\theta}|}$
        & $\frac{|\mathcal{V}_{\cap}|}{|\mathcal{V}_{\mathrm{T}}|}$
        & $J_{\text{vocab}}$ \\
        \midrule
        Qwen$\rightarrow$Llama
        & \textcolor{MutedTableText}{151,665}
        & \textcolor{MutedTableText}{128,256}
        & \textcolor{MutedTableText}{109,567}
        & 85.43 & 72.24 & 64.32 \\
        Granite$\rightarrow$Phi
        & \textcolor{MutedTableText}{100,352}
        & \textcolor{MutedTableText}{200,029}
        & \textcolor{MutedTableText}{85,034}
        & 42.51 & 84.74 & 39.49 \\
        Granite$\rightarrow$Qwen
        & \textcolor{MutedTableText}{100,352}
        & \textcolor{MutedTableText}{151,665}
        & \textcolor{MutedTableText}{99,163}
        & 65.38 & 98.82 & 64.87 \\
        \bottomrule
    \end{tabular}
    \end{center}
\end{table}

For the probability-mass probe, each student before distillation generates one response for each prompt in the complete, unfiltered 20k prompts, using the rollout sampling and generation-length settings in Table~\ref{tab:training-configuration} for all three model pairs. The training prompt length filter of 2,048 student tokens is not applied to this probe. For scoring, we exclude empty responses, responses whose teacher and student tokenizations yield different response byte streams, responses without scorable strict positions, and responses whose full encoded input exceeds either model's context limit or an enabled analysis sequence-length cap. We measure both models' retained probability mass on the same student-selected top-$k$ subset from their original full-vocabulary probabilities and average over the remaining scorable strict positions.

\begin{table}[t]
    \caption{Mean retained probability mass (\%) at strict positions before distillation. ``Shared'' denotes the full shared vocabulary, and bold values mark $k=16$.}
    \label{tab:mass}

    \begin{center}
    \begin{tabular}{lrrrrrr}
        \toprule
        \multirow{2}{*}{\textbf{Subset}}
        & \multicolumn{2}{c}{\textbf{Qwen}$\rightarrow$\textbf{Llama}}
        & \multicolumn{2}{c}{\textbf{Granite}$\rightarrow$\textbf{Phi}}
        & \multicolumn{2}{c}{\textbf{Granite}$\rightarrow$\textbf{Qwen}} \\
        \cmidrule(lr){2-3}\cmidrule(lr){4-5}\cmidrule(lr){6-7}
        & \textbf{Student} & \textbf{Teacher}
        & \textbf{Student} & \textbf{Teacher}
        & \textbf{Student} & \textbf{Teacher} \\
        \midrule
        $k=4$ & 91.74 & 89.81 & 90.08 & 88.12 & 96.58 & 94.53 \\
        $k=8$ & 93.60 & 92.43 & 92.90 & 91.43 & 97.81 & 96.48 \\
        $k=16$ & \textbf{94.55} & \textbf{93.92} & \textbf{94.68} & \textbf{93.54} & \textbf{98.44} & \textbf{97.67} \\
        $k=32$ & 95.06 & 94.77 & 95.89 & 95.03 & 98.82 & 98.37 \\
        $k=64$ & 95.35 & 95.29 & 96.75 & 96.14 & 99.07 & 98.82 \\
        $k=128$ & 95.54 & 95.63 & 97.38 & 96.98 & 99.26 & 99.14 \\
        \midrule
        Shared & 99.72 & 99.73 & 98.99 & 99.69 & 99.81 & 99.90 \\
        \bottomrule
    \end{tabular}
    \end{center}
\end{table}

Gradient diagnostics follow Section~\ref{sec:gradient-analysis}, using checkpoints from each model pair's Strict full run with responses to the same 512 prompts at each checkpoint. The strict component uses reverse KL on the full shared vocabulary, and the mismatch component uses squared differences between span log-probabilities. Replay matches the training batch settings. We report medians across accumulated replay units, using all trainable student parameters.

Responses are read from saved rollouts and held fixed during differentiation. We measure the raw accumulated parameter gradients, without gradient clipping, optimizer preconditioning, or parameter updates. The split-half reference partitions the strict positions and paired EOS positions within each batch and accumulates the two subsets' gradients separately over the replay unit.

In accumulated replay, samples are excluded if they have neither response content nor an appended EOS, if a mismatch group's teacher and student byte streams differ, or if either encoded input exceeds its model's context limit or an enabled sequence-length cap. Length-truncated responses are retained without an appended EOS. Provided the analysis-unit limit has not been reached, a nonempty final unit is analyzed even when it contains fewer samples than the target batch size. Its losses use its actual token and batch counts, and its finite statistics are included in the summary.

\begin{table}[t]
    \caption{Gradient diagnostics at checkpoints from training with only the strict loss. $c_{\text{mis}}$ and $c_{\text{ctrl}}$ denote the mismatch--strict and split-half strict gradient cosines, respectively. $\rho$ is the unweighted span-to-strict norm ratio. Values are rounded to three decimals; figure growth factors use the unrounded medians.}
    \label{tab:gradient-geometry}

    \begin{center}
    \begin{tabular}{lrrrrrrrrr}
        \toprule
        \multirow{2}{*}{\textbf{Step}}
        & \multicolumn{3}{c}{\textbf{Qwen}$\rightarrow$\textbf{Llama}}
        & \multicolumn{3}{c}{\textbf{Granite}$\rightarrow$\textbf{Phi}}
        & \multicolumn{3}{c}{\textbf{Granite}$\rightarrow$\textbf{Qwen}} \\
        \cmidrule(lr){2-4}\cmidrule(lr){5-7}\cmidrule(lr){8-10}
        & $c_{\mathrm{mis}}$ & $c_{\mathrm{ctrl}}$ & $\rho$
        & $c_{\mathrm{mis}}$ & $c_{\mathrm{ctrl}}$ & $\rho$
        & $c_{\mathrm{mis}}$ & $c_{\mathrm{ctrl}}$ & $\rho$ \\
        \midrule
        0 & $0.030$ & $0.956$ & $0.294$ & $0.066$ & $0.986$ & $0.008$ & ${-}0.360$ & $0.977$ & $0.015$ \\
        20 & $0.058$ & $0.671$ & $1.231$ & $0.008$ & $0.955$ & $0.021$ & ${-}0.129$ & $0.581$ & $0.113$ \\
        60 & ${-}0.011$ & $0.296$ & $1.877$ & $0.044$ & $0.733$ & $0.100$ & $0.002$ & $0.049$ & $0.222$ \\
        100 & ${-}0.021$ & $0.208$ & $1.942$ & ${-}0.004$ & $0.246$ & $0.233$ & ${-}0.001$ & $0.076$ & $0.269$ \\
        \bottomrule
    \end{tabular}
    \end{center}
\end{table}

\section{Complete Downstream Results}
\label{app:complete-results}

All scores are accuracies in percent, using mean@32 for mathematics and mean@8 for code. Distilled model scores are evaluated at step 100. Math, Code, and Full follow the settings in Appendix~\ref{app:experimental-details}.

\begin{table}[ht]
    \caption{Benchmark-level scores of the undistilled students and frozen teachers for the three main model pairs.}
    \label{tab:reference-model-scores}

    \begin{center}
    \begin{tabular}{lrrrrrr}
        \toprule
        \multirow{2}{*}{\textbf{Benchmark}}
        & \multicolumn{2}{c}{\textbf{Qwen}$\rightarrow$\textbf{Llama}}
        & \multicolumn{2}{c}{\textbf{Granite}$\rightarrow$\textbf{Phi}}
        & \multicolumn{2}{c}{\textbf{Granite}$\rightarrow$\textbf{Qwen}} \\
        \cmidrule(lr){2-3}\cmidrule(lr){4-5}\cmidrule(lr){6-7}
        & \textbf{Base} & \textbf{Teacher} & \textbf{Base} & \textbf{Teacher} & \textbf{Base} & \textbf{Teacher} \\
        \midrule
        MATH500 & 34.32 & 74.50 & 62.94 & 80.81 & 48.60 & 80.81 \\
        GSM8K & 63.44 & 89.93 & 86.07 & 91.51 & 67.65 & 91.51 \\
        AIME-2024 & 2.29 & 11.88 & 6.35 & 22.71 & 5.10 & 22.71 \\
        AIME-2025 & 0.31 & 7.50 & 3.85 & 18.65 & 1.77 & 18.65 \\
        AIME-2026 & 0.52 & 7.50 & 4.69 & 12.81 & 2.81 & 12.81 \\
        AMC23 & 15.55 & 50.94 & 35.86 & 64.53 & 28.05 & 64.53 \\
        Minerva-Math & 11.82 & 35.09 & 31.69 & 39.76 & 18.29 & 39.76 \\
        \midrule
        HumanEval & 44.89 & 72.94 & 55.34 & 74.54 & 44.89 & 74.54 \\
        MBPP & 53.27 & 77.71 & 49.34 & 82.97 & 45.77 & 82.97 \\
        LiveCodeBench & 8.63 & 24.64 & 15.45 & 32.46 & 12.86 & 32.46 \\
        \midrule
        Math & 18.32 & 39.62 & 33.06 & 47.25 & 24.61 & 47.25 \\
        Code & 35.60 & 58.43 & 40.04 & 63.32 & 34.51 & 63.32 \\
        Full & 26.96 & 49.03 & 36.55 & 55.29 & 29.56 & 55.29 \\
        \bottomrule
    \end{tabular}
    \end{center}
\end{table}

\begin{table}[ht]
    \caption{Benchmark-level comparison for Qwen2.5-7B-Instruct$\rightarrow$Llama-3.2-3B-Instruct.}
    \label{tab:detailed-main-qwen-llama}

    \begin{center}
    \begin{tabular}{lrrrrrrr}
        \toprule
        \multirow{2}{*}{\textbf{Benchmark}}
        & \multirow{2}{*}{\textbf{ULD}}
        & \multirow{2}{*}{\shortstack{\textbf{Extended}\\\textbf{ULD}}}
        & \multirow{2}{*}{\textbf{GOLD}}
        & \multirow{2}{*}{\textbf{SimCT}}
        & \multicolumn{3}{c}{\textbf{Strict}} \\
        \cmidrule(lr){6-8}
        & & & & & \textbf{full} & \textbf{top-16} & \textbf{top-128} \\
        \midrule
        MATH500 & 40.73 & 47.33 & 41.70 & 50.55 & 52.74 & 52.92 & 52.08 \\
        GSM8K & 62.11 & 63.96 & 70.00 & 78.20 & 80.21 & 80.26 & 80.08 \\
        AIME-2024 & 5.00 & 5.73 & 3.33 & 3.65 & 4.69 & 5.00 & 5.10 \\
        AIME-2025 & 0.52 & 1.15 & 0.62 & 1.56 & 1.67 & 1.88 & 1.98 \\
        AIME-2026 & 0.94 & 0.73 & 0.00 & 0.73 & 0.94 & 1.25 & 0.52 \\
        AMC23 & 19.92 & 24.77 & 18.67 & 25.00 & 27.66 & 28.67 & 26.64 \\
        Minerva-Math & 14.34 & 17.37 & 14.15 & 17.47 & 18.31 & 18.07 & 17.80 \\
        \midrule
        HumanEval & 48.09 & 51.98 & 43.22 & 50.91 & 52.74 & 51.52 & 52.90 \\
        MBPP & 55.06 & 52.98 & 45.70 & 49.67 & 51.46 & 50.86 & 49.77 \\
        LiveCodeBench & 10.07 & 12.83 & 10.38 & 13.01 & 13.13 & 12.89 & 12.64 \\
        \midrule
        Math & 20.51 & 23.01 & 21.21 & 25.31 & 26.60 & \textbf{26.86} & 26.31 \\
        Code & 37.74 & \textbf{39.26} & 33.10 & 37.86 & 39.11 & 38.42 & 38.44 \\
        Full & 29.12 & 31.13 & 27.16 & 31.59 & \textbf{32.86} & 32.64 & 32.38 \\
        \bottomrule
    \end{tabular}
    \end{center}
\end{table}

\begin{table}[ht]
    \caption{Benchmark-level comparison for Granite-4.1-8B$\rightarrow$Phi-4-mini-instruct.}
    \label{tab:detailed-main-granite-phi}

    \begin{center}
    \begin{tabular}{lrrrrrrr}
        \toprule
        \multirow{2}{*}{\textbf{Benchmark}}
        & \multirow{2}{*}{\textbf{ULD}}
        & \multirow{2}{*}{\shortstack{\textbf{Extended}\\\textbf{ULD}}}
        & \multirow{2}{*}{\textbf{GOLD}}
        & \multirow{2}{*}{\textbf{SimCT}}
        & \multicolumn{3}{c}{\textbf{Strict}} \\
        \cmidrule(lr){6-8}
        & & & & & \textbf{full} & \textbf{top-16} & \textbf{top-128} \\
        \midrule
        MATH500 & 68.20 & 68.11 & 69.06 & 68.86 & 70.55 & 70.10 & 70.26 \\
        GSM8K & 86.00 & 86.25 & 87.06 & 86.95 & 87.53 & 87.56 & 87.43 \\
        AIME-2024 & 8.85 & 8.33 & 9.48 & 10.73 & 11.25 & 10.83 & 11.46 \\
        AIME-2025 & 3.44 & 5.21 & 6.56 & 5.83 & 6.46 & 7.19 & 6.98 \\
        AIME-2026 & 5.52 & 6.15 & 6.25 & 6.88 & 7.81 & 7.29 & 6.77 \\
        AMC23 & 43.05 & 40.47 & 42.03 & 44.69 & 45.86 & 45.00 & 45.16 \\
        Minerva-Math & 31.65 & 32.46 & 33.08 & 32.74 & 32.96 & 32.65 & 32.85 \\
        \midrule
        HumanEval & 54.42 & 53.12 & 67.07 & 65.17 & 66.01 & 65.02 & 67.00 \\
        MBPP & 52.65 & 52.38 & 53.97 & 54.17 & 54.93 & 55.79 & 55.75 \\
        LiveCodeBench & 18.95 & 18.76 & 20.83 & 20.76 & 21.53 & 21.07 & 20.70 \\
        \midrule
        Math & 35.24 & 35.28 & 36.22 & 36.67 & \textbf{37.49} & 37.23 & 37.27 \\
        Code & 42.01 & 41.42 & 47.29 & 46.70 & 47.49 & 47.29 & \textbf{47.82} \\
        Full & 38.63 & 38.35 & 41.75 & 41.68 & 42.49 & 42.26 & \textbf{42.54} \\
        \bottomrule
    \end{tabular}
    \end{center}
\end{table}

\begin{table}[ht]
    \caption{Benchmark-level comparison for Granite-4.1-8B$\rightarrow$Qwen2.5-7B-Base.}
    \label{tab:detailed-main-granite-qwen}

    \begin{center}
    \begin{tabular}{lrrrrrrr}
        \toprule
        \multirow{2}{*}{\textbf{Benchmark}}
        & \multirow{2}{*}{\textbf{ULD}}
        & \multirow{2}{*}{\shortstack{\textbf{Extended}\\\textbf{ULD}}}
        & \multirow{2}{*}{\textbf{GOLD}}
        & \multirow{2}{*}{\textbf{SimCT}}
        & \multicolumn{3}{c}{\textbf{Strict}} \\
        \cmidrule(lr){6-8}
        & & & & & \textbf{full} & \textbf{top-16} & \textbf{top-128} \\
        \midrule
        MATH500 & 65.84 & 66.16 & 73.31 & 73.73 & 74.91 & 75.25 & 75.31 \\
        GSM8K & 85.63 & 83.97 & 88.03 & 87.75 & 88.64 & 88.81 & 88.87 \\
        AIME-2024 & 7.60 & 8.23 & 10.94 & 11.67 & 14.37 & 14.17 & 14.17 \\
        AIME-2025 & 3.85 & 4.38 & 8.33 & 9.06 & 9.27 & 8.75 & 9.69 \\
        AIME-2026 & 6.98 & 5.31 & 7.29 & 5.73 & 6.25 & 6.46 & 6.46 \\
        AMC23 & 38.67 & 39.30 & 46.88 & 49.69 & 52.11 & 46.95 & 50.00 \\
        Minerva-Math & 24.98 & 24.61 & 33.04 & 32.86 & 33.40 & 33.08 & 33.17 \\
        \midrule
        HumanEval & 34.15 & 51.07 & 67.84 & 68.60 & 68.67 & 69.74 & 69.21 \\
        MBPP & 20.37 & 35.02 & 71.69 & 71.20 & 71.56 & 71.06 & 73.21 \\
        LiveCodeBench & 8.10 & 9.91 & 24.99 & 20.88 & 24.29 & 24.37 & 24.45 \\
        \midrule
        Math & 33.36 & 33.14 & 38.26 & 38.64 & \textbf{39.85} & 39.07 & 39.67 \\
        Code & 20.87 & 32.00 & 54.84 & 53.56 & 54.84 & 55.06 & \textbf{55.62} \\
        Full & 27.12 & 32.57 & 46.55 & 46.10 & 47.35 & 47.06 & \textbf{47.65} \\
        \bottomrule
    \end{tabular}
    \end{center}
\end{table}

\begin{table}[ht]
    \caption{Benchmark-level span-MSE weight sweep for Qwen2.5-7B-Instruct$\rightarrow$Llama-3.2-3B-Instruct.}
    \label{tab:detailed-span-qwen-llama}

    \begin{center}
    \begin{tabular}{lrrrrrrr}
        \toprule
        \multirow{2}{*}{\textbf{Benchmark}} & \multicolumn{7}{c}{\textbf{Span-MSE weight} $\lambda$} \\
        \cmidrule(lr){2-8}
        & \textbf{0} & \textbf{0.25} & \textbf{0.5} & \textbf{0.75} & \textbf{1} & \textbf{1.25} & \textbf{1.5} \\
        \midrule
        MATH500 & 52.74 & 52.38 & 51.21 & 51.19 & 50.50 & 50.18 & 49.95 \\
        GSM8K & 80.21 & 79.96 & 79.35 & 79.65 & 79.11 & 78.94 & 78.75 \\
        AIME-2024 & 4.69 & 5.10 & 5.21 & 5.21 & 4.58 & 4.06 & 5.62 \\
        AIME-2025 & 1.67 & 1.77 & 1.56 & 1.56 & 1.35 & 1.25 & 1.25 \\
        AIME-2026 & 0.94 & 0.42 & 0.62 & 0.31 & 0.42 & 0.52 & 0.21 \\
        AMC23 & 27.66 & 27.81 & 26.02 & 26.72 & 25.47 & 25.78 & 24.92 \\
        Minerva-Math & 18.31 & 17.77 & 17.52 & 17.42 & 17.07 & 17.04 & 17.39 \\
        \midrule
        HumanEval & 52.74 & 52.36 & 51.98 & 52.06 & 50.08 & 49.92 & 50.84 \\
        MBPP & 51.46 & 50.46 & 52.08 & 51.79 & 51.12 & 51.39 & 51.39 \\
        LiveCodeBench & 13.13 & 12.64 & 12.29 & 12.95 & 12.44 & 12.48 & 12.50 \\
        \midrule
        Math & \textbf{26.60} & 26.46 & 25.93 & 26.01 & 25.50 & 25.40 & 25.44 \\
        Code & \textbf{39.11} & 38.49 & 38.78 & 38.93 & 37.88 & 37.93 & 38.24 \\
        Full & \textbf{32.86} & 32.47 & 32.36 & 32.47 & 31.69 & 31.66 & 31.84 \\
        \bottomrule
    \end{tabular}
    \end{center}
\end{table}

\begin{table}[ht]
    \caption{Benchmark-level span-MSE weight sweep for Granite-4.1-8B$\rightarrow$Phi-4-mini-instruct.}
    \label{tab:detailed-span-granite-phi}

    \begin{center}
    \begin{tabular}{lrrrrrrr}
        \toprule
        \multirow{2}{*}{\textbf{Benchmark}} & \multicolumn{7}{c}{\textbf{Span-MSE weight} $\lambda$} \\
        \cmidrule(lr){2-8}
        & \textbf{0} & \textbf{0.25} & \textbf{0.5} & \textbf{0.75} & \textbf{1} & \textbf{1.25} & \textbf{1.5} \\
        \midrule
        MATH500 & 70.55 & 71.04 & 70.01 & 70.44 & 69.59 & 70.47 & 70.08 \\
        GSM8K & 87.53 & 87.64 & 87.31 & 87.57 & 87.09 & 87.31 & 87.46 \\
        AIME-2024 & 11.25 & 10.31 & 10.42 & 10.31 & 9.69 & 10.31 & 10.00 \\
        AIME-2025 & 6.46 & 6.67 & 6.46 & 6.77 & 7.08 & 6.25 & 6.46 \\
        AIME-2026 & 7.81 & 5.94 & 6.77 & 7.19 & 6.88 & 6.77 & 6.35 \\
        AMC23 & 45.86 & 43.36 & 44.06 & 46.25 & 44.77 & 44.92 & 43.36 \\
        Minerva-Math & 32.96 & 32.96 & 32.73 & 32.12 & 31.99 & 33.50 & 32.12 \\
        \midrule
        HumanEval & 66.01 & 65.09 & 63.64 & 64.41 & 64.48 & 63.64 & 63.19 \\
        MBPP & 54.93 & 54.79 & 55.56 & 55.89 & 53.21 & 56.55 & 56.22 \\
        LiveCodeBench & 21.53 & 20.84 & 21.21 & 20.51 & 20.63 & 20.68 & 21.11 \\
        \midrule
        Math & \textbf{37.49} & 36.85 & 36.82 & 37.24 & 36.73 & 37.08 & 36.55 \\
        Code & \textbf{47.49} & 46.91 & 46.80 & 46.94 & 46.11 & 46.96 & 46.84 \\
        Full & \textbf{42.49} & 41.88 & 41.81 & 42.09 & 41.42 & 42.02 & 41.69 \\
        \bottomrule
    \end{tabular}
    \end{center}
\end{table}

\begin{table}[ht]
    \caption{Benchmark-level span-MSE weight sweep for Granite-4.1-8B$\rightarrow$Qwen2.5-7B-Base.}
    \label{tab:detailed-span-granite-qwen}

    \begin{center}
    \begin{tabular}{lrrrrrrr}
        \toprule
        \multirow{2}{*}{\textbf{Benchmark}} & \multicolumn{7}{c}{\textbf{Span-MSE weight} $\lambda$} \\
        \cmidrule(lr){2-8}
        & \textbf{0} & \textbf{0.25} & \textbf{0.5} & \textbf{0.75} & \textbf{1} & \textbf{1.25} & \textbf{1.5} \\
        \midrule
        MATH500 & 74.91 & 75.45 & 74.97 & 74.84 & 74.54 & 74.72 & 74.81 \\
        GSM8K & 88.64 & 88.79 & 88.75 & 88.77 & 88.83 & 88.95 & 88.77 \\
        AIME-2024 & 14.37 & 15.42 & 13.02 & 14.27 & 12.71 & 12.19 & 12.92 \\
        AIME-2025 & 9.27 & 8.02 & 9.27 & 9.69 & 8.23 & 8.02 & 8.02 \\
        AIME-2026 & 6.25 & 6.15 & 5.94 & 6.15 & 5.83 & 5.83 & 5.52 \\
        AMC23 & 52.11 & 49.45 & 49.45 & 50.62 & 48.44 & 49.06 & 49.30 \\
        Minerva-Math & 33.40 & 32.97 & 33.94 & 33.48 & 33.66 & 33.85 & 33.43 \\
        \midrule
        HumanEval & 68.67 & 67.99 & 68.22 & 67.45 & 67.38 & 67.99 & 67.99 \\
        MBPP & 71.56 & 71.23 & 71.79 & 70.57 & 70.90 & 71.76 & 70.57 \\
        LiveCodeBench & 24.29 & 24.55 & 24.44 & 23.86 & 24.10 & 24.15 & 23.86 \\
        \midrule
        Math & \textbf{39.85} & 39.46 & 39.33 & 39.69 & 38.89 & 38.95 & 38.97 \\
        Code & \textbf{54.84} & 54.59 & 54.82 & 53.96 & 54.13 & 54.63 & 54.14 \\
        Full & \textbf{47.35} & 47.03 & 47.08 & 46.82 & 46.51 & 46.79 & 46.55 \\
        \bottomrule
    \end{tabular}
    \end{center}
\end{table}

\section{Additional Results with a Larger Teacher}
\label{app:qwen3-granite}

We evaluate strict supervision with Qwen3-235B-A22B-Instruct-2507$\rightarrow$Granite-4.1-8B, using Strict top-16 from Section~\ref{sec:method}. Tables~\ref{tab:qwen3-granite-reference} and~\ref{tab:detailed-qwen3-granite} report performance, with Base and Teacher denoting the undistilled student and frozen teacher. The mathematics domain adds HMMT 2025 February and November \citep{dekoninck2026beyond} to the seven benchmarks in Appendix~\ref{app:experimental-details}.

\begin{table}[ht]
    \caption{Benchmark-level scores of the undistilled student and frozen teacher for Qwen3-235B-A22B-Instruct-2507$\rightarrow$Granite-4.1-8B.}
    \label{tab:qwen3-granite-reference}

    \begin{center}
    \begin{tabular}{lrr}
        \toprule
        \textbf{Benchmark} & \textbf{Base} & \textbf{Teacher} \\
        \midrule
        MATH500 & 80.81 & 95.88 \\
        GSM8K & 91.51 & 94.24 \\
        AIME-2024 & 22.71 & 60.94 \\
        AIME-2025 & 18.65 & 51.98 \\
        AIME-2026 & 12.81 & 65.52 \\
        AMC23 & 64.53 & 90.00 \\
        Minerva-Math & 39.76 & 48.10 \\
        HMMT 2025 February & 10.00 & 41.15 \\
        HMMT 2025 November & 12.92 & 46.15 \\
        \midrule
        HumanEval & 74.54 & 87.12 \\
        MBPP & 82.97 & 94.54 \\
        LiveCodeBench & 32.46 & 65.38 \\
        \midrule
        Math & 39.30 & 66.00 \\
        Code & 63.32 & 82.35 \\
        Full & 51.31 & 74.17 \\
        \bottomrule
    \end{tabular}
    \end{center}
\end{table}

\begin{table}[ht]
    \caption{Benchmark-level distillation results for Qwen3-235B-A22B-Instruct-2507$\rightarrow$Granite-4.1-8B. Bold marks the best Math, Code, and Full averages.}
    \label{tab:detailed-qwen3-granite}

    \begin{center}
    \begin{tabular}{lrrrrr}
        \toprule
        \textbf{Benchmark} & \textbf{ULD} & \shortstack{\textbf{Extended}\\\textbf{ULD}} & \textbf{GOLD} & \textbf{SimCT} & \shortstack{\textbf{Strict}\\\textbf{top-16}} \\
        \midrule
        MATH500 & 81.31 & 87.29 & 85.91 & 86.08 & 89.75 \\
        GSM8K & 91.59 & 93.06 & 93.12 & 93.15 & 93.69 \\
        AIME-2024 & 23.23 & 37.19 & 35.52 & 37.40 & 43.02 \\
        AIME-2025 & 21.15 & 26.98 & 27.19 & 29.17 & 35.10 \\
        AIME-2026 & 13.33 & 30.10 & 23.02 & 27.29 & 38.44 \\
        AMC23 & 67.34 & 72.66 & 69.30 & 74.22 & 76.80 \\
        Minerva-Math & 40.44 & 41.22 & 43.35 & 41.58 & 44.11 \\
        HMMT 2025 February & 11.56 & 12.60 & 12.29 & 10.94 & 16.67 \\
        HMMT 2025 November & 13.12 & 17.19 & 17.81 & 18.75 & 24.06 \\
        \midrule
        HumanEval & 76.98 & 65.40 & 76.45 & 82.32 & 91.54 \\
        MBPP & 84.09 & 69.11 & 76.03 & 78.90 & 85.55 \\
        LiveCodeBench & 32.48 & 33.01 & 35.73 & 36.74 & 43.76 \\
        \midrule
        Math & 40.34 & 46.48 & 45.28 & 46.51 & \textbf{51.29} \\
        Code & 64.52 & 55.84 & 62.74 & 65.99 & \textbf{73.62} \\
        Full & 52.43 & 51.16 & 54.01 & 56.25 & \textbf{62.46} \\
        \bottomrule
    \end{tabular}
    \end{center}
\end{table}

\section{Additional Results on ALFWorld}
\label{app:alfworld}

We evaluate Cross-Tokenizer On-Policy Distillation on ALFWorld \citep{shridhar2021alfworld}, using Qwen3-4B-Instruct-2507$\rightarrow$Llama-3.2-3B-Instruct and the ALFWorld training split. Table~\ref{tab:alfworld-results} compares Strict top-16 from Section~\ref{sec:method} with ULD, Extended ULD, GOLD, and SimCT. Base denotes the student before distillation, and Teacher denotes the frozen teacher. We evaluate on the official seen and unseen splits, using 128 episodes per evaluation run and a limit of 50 interaction turns per episode. Scores are averaged over three evaluations. SR is the success rate in percent, and Turns denotes the average number of interaction turns across all episodes, including successes and failures. Overall is the arithmetic mean of the seen and unseen scores for each metric.

\begin{table}[ht]
    \caption{ALFWorld results for Qwen3-4B-Instruct-2507$\rightarrow$Llama-3.2-3B-Instruct. Bold marks the highest SR and lowest Turns among distilled models.}
    \label{tab:alfworld-results}

    \begin{center}
    \begin{tabular}{lrrrrrr}
        \toprule
        \multirow{2}{*}{\textbf{Method}} & \multicolumn{2}{c}{\textbf{Seen}} & \multicolumn{2}{c}{\textbf{Unseen}} & \multicolumn{2}{c}{\textbf{Overall}} \\
        \cmidrule(lr){2-3}\cmidrule(lr){4-5}\cmidrule(lr){6-7}
        & \textbf{SR} $\uparrow$ & \textbf{Turns} $\downarrow$ & \textbf{SR} $\uparrow$ & \textbf{Turns} $\downarrow$ & \textbf{SR} $\uparrow$ & \textbf{Turns} $\downarrow$ \\
        \midrule
        Base & 12.50 & 46.60 & 8.85 & 47.28 & 10.68 & 46.94 \\
        Teacher & 29.69 & 41.07 & 29.43 & 41.22 & 29.56 & 41.15 \\
        \midrule
        ULD & 1.56 & 49.61 & 1.30 & 49.74 & 1.43 & 49.68 \\
        Extended ULD & 25.52 & 41.95 & 23.44 & 42.61 & 24.48 & 42.28 \\
        GOLD & 3.65 & 49.30 & 1.56 & 49.58 & 2.61 & 49.44 \\
        SimCT & 25.00 & 42.31 & 32.03 & 39.31 & 28.52 & 40.81 \\
        \midrule
        Strict top-16 & \textbf{26.04} & \textbf{40.73} & \textbf{34.64} & \textbf{38.82} & \textbf{30.34} & \textbf{39.78} \\
        \bottomrule
    \end{tabular}
    \end{center}
\end{table}

\end{document}